\documentclass[a4paper]{llncs}

\usepackage[T1]{fontenc}
\usepackage{lmodern}
\usepackage{booktabs}
\usepackage{amsmath}
\usepackage{amssymb}
\usepackage{graphicx}
\graphicspath{{figures/}}
\usepackage{placeins}
\usepackage{algorithm}
\usepackage{algpseudocode}
\usepackage[hidelinks]{hyperref}

\title{Mechanism-Guided Selective Unlearning for RLVR-Induced Reasoning}
\titlerunning{Mechanism-Guided Selective Unlearning}

\author{Chenyu Zhou\inst{1}\textsuperscript{*\dag} \and Qiliang Jiang\inst{2}\textsuperscript{*} \and Shuning Wu\inst{3} \and Xu Zhou\inst{3}\textsuperscript{\dag}}
\authorrunning{C. Zhou et al.}
\institute{
School of Engineering, Institute of Science Tokyo, Japan\\
\email{zhou.c.76d6@m.isct.ac.jp}
\and
College of Control Science and Engineering, Zhejiang University, China\\
\email{jiangqiliang@zju.edu.cn}
\and
Department of Electrical and Computer Engineering, National University of Singapore, Singapore\\
\email{shuningwu@u.nus.edu, zhouxu\_nus@u.nus.edu}\\[0.5ex]
\textsuperscript{*}Equal contribution.
\textsuperscript{\dag}Corresponding authors.
}

\begin{document}
\maketitle

\begin{abstract}
We propose MAST (Mechanism-Aligned Selective Targeting), a mechanism-guided method for unlearning RLVR-induced reasoning with substantially lower collateral damage than standard full-parameter updates. In matched SFT/RLVR checkpoints on Qwen2.5-Math-1.5B and Qwen3-1.7B-Base, the SFT-to-RLVR increment differs sharply from the SFT update in token-level delta-log-probability, and full-parameter gradient ascent forgets only by damaging retain MATH and GSM8K. MAST ranks attention-projection tensors by off-principal energy, update magnitude, and forget-gradient coupling magnitude, then updates only the top-ranked subset. On the primary model, MAST induces statistically significant target forgetting (MATH forget 45/150 to 37/150; McNemar $p=0.0078$) while preserving GSM8K ($+0.8$ pp) and MATH retain ($-0.5$ pp). The advantage reproduces across seeds, NPO/SimNPO objectives, and Qwen3, where MAST preserves GSM8K while full-parameter unlearning collapses it.
\end{abstract}

\section{Introduction}

LLM unlearning asks a model to suppress a target behavior or knowledge cluster while preserving non-target capabilities. Existing methods often apply local updates to a post-trained model, such as gradient ascent on forget examples, retain-aware variants, negative preference objectives, or representation control. These tools are commonly evaluated in settings where the target behavior was introduced or emphasized through supervised fine-tuning. That evaluation regime leaves open a practical question: do the same unlearning targets remain appropriate when the behavior was induced by RLVR?

This question matters because RLVR is not merely SFT with a different loss. Recent work suggests that RLVR can shift model behavior through direction-specific probability changes or off-principal update components, rather than through the same high-variance directions emphasized by SFT \cite{rlvr_path_not_taken,rlvr_direction_updates,rlvr_capability_boundary}. If standard unlearning updates implicitly target SFT-like directions, then they can move accuracy numbers while still updating the wrong behavioral subspace or causing avoidable collateral damage.

We study this hypothesis in a controlled mathematical reasoning setting on two model families. Starting from Qwen2.5-Math-1.5B (primary) and Qwen3-1.7B-Base (cross-model), the protocol constructs SFT and RLVR checkpoints from the same base model on matched GSM8K data. It then evaluates forgetting on MATH category splits while measuring non-target MATH and GSM8K performance. We use ``mis-targeting'' in a directional sense: the standard forget update is poorly aligned with the directions RLVR actually moves.

The paper makes four contributions, validated on the primary model and corroborated on a second model family:
\begin{enumerate}
  \item \textbf{Mechanism separation.} Under matched target accuracy, the SFT update and the SFT-to-RLVR increment are distinguishable by token-level delta-log-probability: SFT is a positive likelihood boost on existing continuations, while the RLVR increment is balanced and direction-specific. The same direction reproduces on the second model.
  \item \textbf{No clean operating point for full-parameter unlearning.} Standard full-parameter gradient ascent on the RLVR checkpoint reduces the target metric only by inflicting large collateral damage on both retain MATH and GSM8K. This boundary reproduces on both models and shows that forget accuracy alone is not a sufficient unlearning signal.
  \item \textbf{MAST: a mechanism-guided unlearning method.} We turn the diagnosis into a method, \emph{Mechanism-Aligned Selective Targeting} (MAST), which ranks attention-projection tensors by a mechanism score and restricts the unlearning update to the top-ranked subset (Algorithm~\ref{alg:mast}). MAST achieves meaningful forgetting at substantially lower collateral than full-parameter unlearning; on the primary model the forget drop is statistically significant, and the restricted-versus-full collateral advantage is statistically significant on both models. The mechanism-based ranking---not merely the smaller parameter set---drives the lower collateral: the selected target set is structurally coherent by construction (all query/output projections), the lowest-ranked tensors produce negligible forgetting, and the selected set retains significantly more GSM8K than a same-size random subset at matched forgetting.
  \item \textbf{An evaluation pitfall.} Beyond MAST, we show that forget-accuracy-only and answer-likelihood-only evaluations are each incomplete for reasoning unlearning: an update can lower solution-trajectory likelihood (degrading generated reasoning) while \emph{raising} final-answer likelihood, so the two views disagree (Section~\ref{sec:eval}).
\end{enumerate}

\section{Related Work}

\paragraph{LLM unlearning.}
LLM unlearning methods include gradient ascent, retain-aware objectives, preference-style negative objectives such as NPO and SimNPO, and representation-control methods such as RMU \cite{npo,simnpo,wmdp}. Benchmarks such as TOFU and WMDP emphasize that unlearning must be evaluated jointly on forgetting and retained utility \cite{tofu,wmdp,llm_unlearning_survey}. This paper differs by focusing on a post-training source mismatch: the target behavior is induced by either SFT or RLVR, while the unlearning tool is held fixed.

\paragraph{RLVR and reasoning updates.}
RLVR work on reasoning raises two relevant points. First, capability-boundary analyses argue that RLVR may change sampling efficiency or probability mass over solutions rather than create wholly new capabilities \cite{rlvr_capability_boundary}. Second, geometry and token-direction studies suggest that RLVR updates can live in off-principal or direction-specific components \cite{rlvr_path_not_taken,rlvr_direction_updates,geora}.

\paragraph{Model diffing and side effects.}
Model diffing and sparse-difference tools provide a broader context for localizing fine-tuning effects and predicting side effects \cite{delta_crosscoder,mneme}. Our use of delta log-probability, update geometry, and gradient alignment is similarly diagnostic, but the experimental target is the unlearning mismatch between SFT-induced and RLVR-induced reasoning. Our target-tensor ranking follows the localization-and-specificity paradigm of knowledge-editing methods \cite{rome,memit,localization_editing}, which validate a localized intervention against random-location controls and reverse ablations rather than by edit magnitude alone.

\section{Benchmark and Controls}

\subsection{Model and Training Data}

The primary base model is Qwen2.5-Math-1.5B; the cross-model family (Section~\ref{sec:crossmodel}) is Qwen3-1.7B-Base. Both post-training branches use the same prepared GSM8K matched data: 2048 training examples and 256 evaluation examples. The formal SFT checkpoint and the RLVR checkpoint are evaluated with the same scripts and splits. Training outputs record optimizer steps, tokens seen, seed, and training metrics. Evaluation uses greedy decoding; exact match is the primary metric, with a tolerant regrade as a robustness check.

\begin{table}[!htbp]
\centering
\small
\begin{tabular*}{\textwidth}{@{\extracolsep{\fill}}lrrr@{}}
\toprule
Checkpoint & GSM8K acc. & MATH forget acc. & MATH retain acc. \\
\midrule
Base & 0.4844 & 0.2267 & 0.2900 \\
SFT & 0.7461 & 0.2867 & 0.3950 \\
RLVR & 0.7344 & 0.3000 & 0.3850 \\
\bottomrule
\end{tabular*}
\caption{Matched checkpoint control table (primary model). SFT and RLVR differ by 1.172 percentage points on GSM8K, below the 3 percentage point matching threshold.}
\label{tab:matched-controls}
\end{table}

\subsection{Checkpoint Matching}

The SFT/RLVR comparison uses a strict target-accuracy match. The SFT checkpoint obtains 191/256 on GSM8K, while the SFT-initialized RLVR checkpoint obtains 188/256, a 1.172 percentage point gap. MATH forget/retain are also close: SFT obtains 43/150 and 79/200, while RLVR obtains 45/150 and 77/200. This makes downstream differences less likely to be explained by raw target ability mismatch.

\subsection{Forget and Retain Sets}
\label{sec:forget-retain}

The forget set uses MATH category splits: Counting and Probability, Geometry, and Number Theory ($N=150$). The retain set uses non-target MATH categories ($N=200$) and GSM8K holdout performance. Exact match is the primary metric; tolerant regrading is a robustness check for numeric equivalence, fractions, tuples, degree notation, and boxed forms. In the primary regrade audit, applying the tolerant regrade uniformly to base, SFT, RLVR, MAST (GA), full GA, and MAST (NPO) leaves every forget split unchanged (0 examples) and moves each retain split by at most 2 examples, confirming that the reported effects are not grading artifacts.

\section{Standard Unlearning Has No Clean Operating Point}

We first evaluate standard SFT-era methods on the matched RLVR checkpoint: gradient ascent on forget examples (GA) and forget gradient ascent plus retain cross-entropy (GA+retain). Across a strength sweep, standard full-parameter GA has no clean operating window. Weak settings barely move the target metric, while stronger settings collapse evaluation accuracy toward zero (Figure~\ref{fig:pareto}).

Even at a controlled mid strength (50 steps, $\eta=5{\times}10^{-6}$), full-parameter GA achieves target forgetting only by destroying utility: GSM8K falls to 172/256, MATH forget to 22/150, and MATH retain to 56/200, for drops of 6.25, 15.33, and 10.5 pp. The full-parameter update reaches a large forget drop, but it removes more than ten points of MATH retain and over six points of GSM8K. Forget accuracy alone therefore overstates unlearning quality: any method can drive the forget metric down if it is allowed to damage everything else. The rest of the paper uses this boundary as the reference against which a localized, direction-aware update must improve. A steps-matched strength sweep (Figure~\ref{fig:pareto}) makes the boundary visual: full-parameter GA has at most one or two usable strengths before catastrophic collapse---on the primary model it forgets meaningfully only at $\eta=5{\times}10^{-6}$, then collapses at $\eta=7{\times}10^{-6}$ (GSM8K 13/256); on Qwen3 only $\eta=5{\times}10^{-6}$ is usable, collapsing by $\eta=8{\times}10^{-6}$ (GSM8K 40/256)---whereas MAST traces a low-collateral Pareto frontier of several usable operating points. No full-parameter strength achieves meaningful forgetting ($\geq$5pp) at low collateral: $\eta=3{\times}10^{-6}$ barely forgets (2pp), $\eta=5{\times}10^{-6}$ forgets but sheds six to ten points of utility, and $\eta\geq7{\times}10^{-6}$ collapses.

Retain regularization does not by itself resolve the boundary. A retain-aware full-parameter objective---forget gradient ascent plus cross-entropy on a retain set---removes essentially no target forgetting at a utility-preserving strength: MATH forget moves only $45\to44$, with GSM8K 187/256 and MATH retain 74/200. The retain term trades the forgetting away rather than opening a clean operating point.

\begin{table}[!htbp]
\centering
\scriptsize
\begin{tabular*}{\textwidth}{@{\extracolsep{\fill}}llrrrl@{}}
\toprule
Model & $\eta$ & GSM8K & forget & retain & note \\
\midrule
\multicolumn{6}{l}{\emph{Primary (Qwen2.5-Math-1.5B), full-parameter GA}} \\
 & $3{\times}10^{-6}$ & 184 & 42 & 80 & weak forget \\
 & $5{\times}10^{-6}$ & 172 & 22 & 56 & boundary point \\
 & $7{\times}10^{-6}$ & 13 & 0 & 7 & collapse \\
 & $\geq1{\times}10^{-5}$ & 0 & 0 & 0 & collapse \\
\multicolumn{6}{l}{\emph{Primary, MAST (top-96)}} \\
 & $3{\times}10^{-6}$ & 188 & 41 & 79 & \\
 & $5{\times}10^{-6}$ & 190 & 37 & 76 & selected \\
 & $7{\times}10^{-6}$ & 185 & 37 & 78 & \\
 & $1{\times}10^{-5}$ & 182 & 28 & 71 & stronger forget \\
 & $1.5{\times}10^{-5}$ & 58 & 2 & 12 & collapse \\
\midrule
\multicolumn{6}{l}{\emph{Qwen3-1.7B-Base, full-parameter GA}} \\
 & $5{\times}10^{-6}$ & 143 & 33 & 46 & boundary point \\
 & $8{\times}10^{-6}$ & 40 & 13 & 19 & collapse \\
 & $1{\times}10^{-5}$ & 9 & 8 & 9 & collapse \\
 & $\geq1.5{\times}10^{-5}$ & 0 & 0 & 0 & collapse \\
\multicolumn{6}{l}{\emph{Qwen3, MAST (top-96)}} \\
 & $5{\times}10^{-6}$ & 171 & 45 & 82 & negligible forgetting \\
 & $8{\times}10^{-6}$ & 176 & 32 & 57 & selected \\
 & $1{\times}10^{-5}$ & 146 & 24 & 43 & \\
 & $1.5{\times}10^{-5}$ & 20 & 8 & 9 & collapse \\
\bottomrule
\end{tabular*}
\caption{Full steps-matched strength frontier underlying Figure~\ref{fig:pareto}. Counts are GSM8K/256, MATH-forget/150, and MATH-retain/200. The Qwen3 rows are the cross-model sweep analyzed in Section~\ref{sec:crossmodel}. At matched forgetting, MAST's collateral is below full-parameter GA throughout the operating region; full-parameter GA collapses after one or two usable strengths on each model.}
\label{tab:frontier}
\end{table}
\FloatBarrier

\section{Mechanism Analysis}
\label{sec:mechanism}

We use three diagnostics to test whether the unlearning mismatch is directional.

\paragraph{Update geometry.}
For each checkpoint, we compute layerwise delta energy between checkpoints and project deltas onto the base-weight top singular subspace. Both updates lie almost entirely off the base-weight principal directions: the fraction of update energy orthogonal to the base tensor's leading singular directions (the off-principal fraction $\rho_t$) is 0.9998 for both, confirming the increment is direction-specific; this geometry provides the selector's provenance for choosing target tensors.

\begin{figure}[!htbp]
\centering
\includegraphics[width=\linewidth]{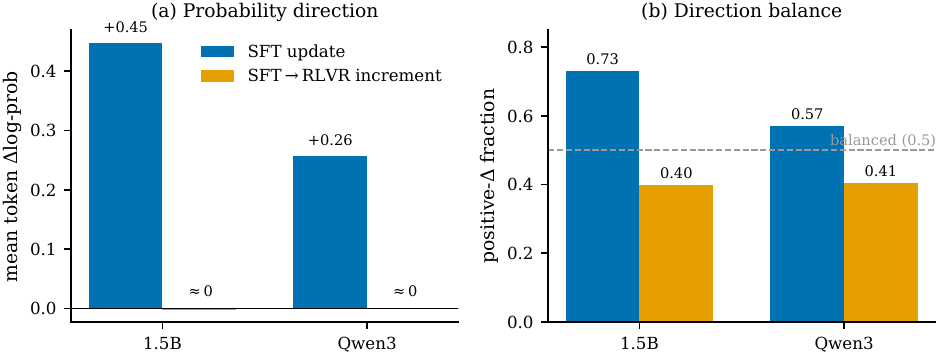}
\caption{Mechanism separation on both model families. (a) The SFT update raises token log-probability (large positive mean $\Delta$log-probability), while the SFT$\to$RLVR increment is near zero. (b) The SFT update is directionally one-sided (positive-$\Delta$ fraction $\gg 0.5$), while the RLVR increment is balanced (near $0.5$). The pattern reproduces on Qwen3-1.7B-Base at smaller magnitude.}
\label{fig:mechanism}
\end{figure}

\paragraph{Token-level delta log-probability.}
The strongest mechanism signal is token-level probability structure (Figure~\ref{fig:mechanism}). SFT produces a positive mean delta-log-probability of 0.4477 and a positive-delta fraction of 0.7296. The SFT-to-RLVR increment has mean delta-log-probability -0.0002 and positive-delta fraction 0.3995. This indicates that the RLVR increment is not simply another SFT-like likelihood boost on the same continuations; it redistributes probability in a more balanced and direction-specific way. The second model reproduces the same direction (Section~\ref{sec:crossmodel}).

\paragraph{Forget-gradient versus checkpoint-delta direction.}
We compute forget-loss gradients on the forget split and compare them with checkpoint deltas. The GA-update cosine with the SFT delta is 0.0110, while the cosine with the SFT-to-RLVR increment is 0.0051. Both are small, and RLVR is lower. We use this as diagnostic evidence that the standard forget gradient weakly tracks the checkpoint directions, especially the incremental RLVR direction.

\begin{table}[t]
\centering
\small
\begin{tabular*}{\textwidth}{@{\extracolsep{\fill}}lrrrr@{}}
\toprule
Checkpoint & Off-princ. frac. & Mean $\Delta$log-prob. & Pos. $\Delta$ frac. & GA cosine \\
\midrule
SFT & 0.9998 & 0.4477 & 0.7296 & 0.0110 \\
RLVR increment & 0.9998 & -0.0002 & 0.3995 & 0.0051 \\
\bottomrule
\end{tabular*}
\caption{Mechanism summary for SFT and the SFT-to-RLVR increment (primary model). Mean delta-log-probability and positive-delta fraction are the primary separation signals.}
\label{tab:mechanism}
\end{table}

\section{MAST: Mechanism-Aligned Selective Targeting}
\label{sec:mast}

Motivated by the diagnosis above --- the SFT-to-RLVR increment is direction-specific and the standard forget gradient tracks it only weakly --- MAST restricts the unlearning update to the attention-projection tensors whose checkpoint change is most off-principal and most strongly coupled to the forget gradient in magnitude. Concretely (Algorithm~\ref{alg:mast}), for each attention tensor $t$ (query/key/value/output across all layers) we score the SFT-to-RLVR increment $\Delta_t = \theta_{R,t}-\theta_{S,t}$ by
\begin{equation}
s_t \;=\; \rho_t \cdot \log\!\left(1+\lVert\Delta_t\rVert_F^2\right)\cdot\bigl(1+\lvert c_t\rvert\bigr),
\label{eq:score}
\end{equation}
where $\mathcal{L}_f$ denotes the forget-set loss, $\rho_t\in[0,1]$ is the fraction of $\Delta_t$'s energy lying off the top singular subspace of the base-model tensor $\theta_{0,t}$ (the off-principal fraction), and $c_t=\cos\bigl(\nabla_{\theta_t}\mathcal{L}_f(\theta_R;D_f),\,\Delta_t\bigr)$ is the cosine between the forget-loss gradient evaluated at the RLVR checkpoint and $\Delta_t$. We use $\lvert c_t\rvert$ because the score localizes tensors whose RLVR increment is coupled to the forget objective; the update direction itself is still set by gradient ascent on $\mathcal{L}_f$. We then update only the top-$k$ tensors by $s_t$, leaving MLP, normalization, and embedding weights frozen; the reported configuration uses $k=96$ of the 112 attention tensors. Section~\ref{sec:ranking} establishes that the ranking, not merely the reduced parameter count, carries the specificity.

\begin{algorithm}[t]
\caption{MAST: Mechanism-Aligned Selective Targeting}
\label{alg:mast}
\begin{algorithmic}[1]
\Require base/SFT/RLVR weights $\theta_0,\theta_S,\theta_R$; forget set $D_f$; selection budget $k$; strength $(\eta,\text{steps})$
\Ensure unlearned weights $\theta'$
\For{each attention-projection tensor $t$ (q/k/v/o, all layers)}
  \State $\Delta_t \gets \theta_{R,t}-\theta_{S,t}$ \Comment{SFT$\to$RLVR increment}
  \State $\rho_t \gets$ off-principal energy of $\Delta_t$ w.r.t.\ top singular subspace of $\theta_{0,t}$, normalized
  \State $c_t \gets \cos\bigl(\nabla_{\theta_t}\mathcal{L}_f(\theta_R;D_f),\,\Delta_t\bigr)$
  \State $s_t \gets \rho_t\cdot\log(1+\lVert\Delta_t\rVert_F^2)\cdot(1+\lvert c_t\rvert)$
\EndFor
\State $T \gets$ indices of the top-$k$ tensors by $s_t$
\State $\theta' \gets$ gradient ascent on $\mathcal{L}_f(D_f)$ at rate $\eta$ for \textit{steps}, updating only $\{\theta_t: t\in T\}$
\State \Return $\theta'$
\end{algorithmic}
\end{algorithm}

Equation~\ref{eq:score} operationalizes the diagnosis of Section~\ref{sec:mechanism}. RLVR redistributes probability through direction-specific changes that, in weight space, lie off the base model's principal directions. MAST scores each attention tensor by how strongly its RLVR increment is off-principal ($\rho_t$), how much the increment changes the tensor ($\lVert\Delta_t\rVert_F^2$), and the magnitude of forget-gradient coupling ($\lvert c_t\rvert$). The score concentrates the unlearning update on the attention projections RLVR reshaped most for reasoning, while leaving MLP knowledge stores and normalization untouched.

\subsection{Selected Operating Point}

\begin{figure}[!htbp]
\centering
\includegraphics[width=\linewidth]{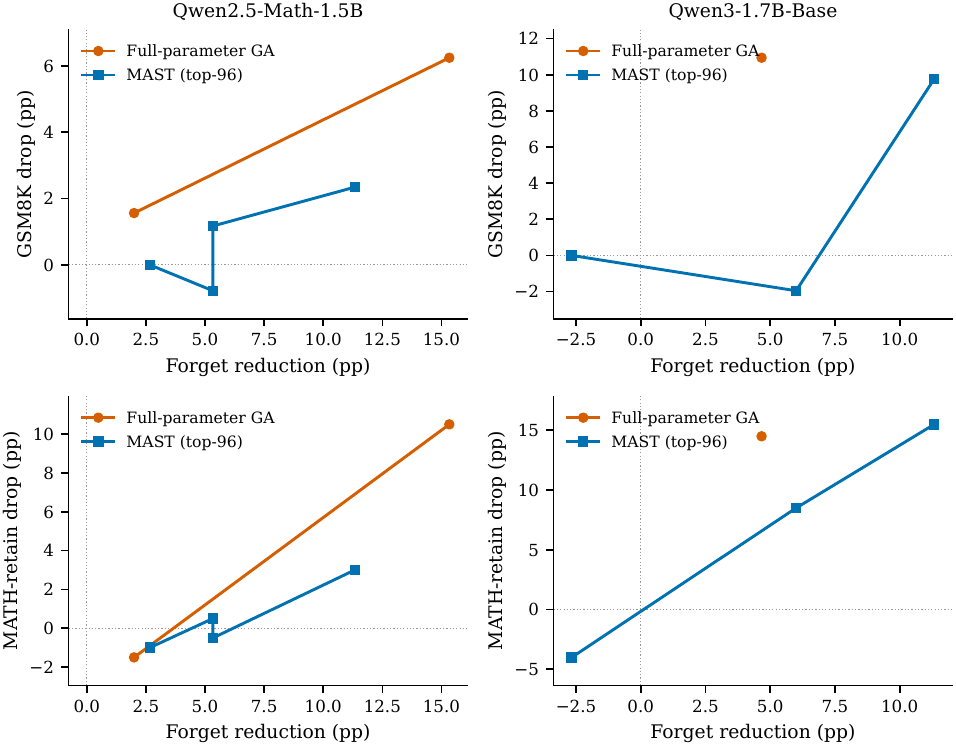}
\caption{Forget--collateral Pareto frontier on both model families (steps-matched GA strength sweep, 50 steps). Each point is one unlearning strength; the $x$-axis is MATH-forget reduction and the $y$-axis is collateral (GSM8K, top row; MATH-retain, bottom row; lower is better). Negative $x$-values mark strengths that increase target accuracy rather than forget. Wherever full-parameter GA forgets meaningfully, its collateral exceeds MAST's at matched forgetting, on both models and both axes. Full-parameter GA has at most one or two usable strengths before catastrophic collapse (primary $\eta=7{\times}10^{-6}$: GSM8K 13/256; Qwen3 $\eta=8{\times}10^{-6}$: GSM8K 40/256) and therefore traces no usable frontier on Qwen3, whereas MAST traces a low-collateral frontier; strengths beyond the last plotted point collapse both methods (full counts in Table~\ref{tab:frontier}).}
\label{fig:pareto}
\end{figure}

On the matched RLVR checkpoint, MAST (GA) at 50 steps and $\eta=5{\times}10^{-6}$ reaches a clean tradeoff that the full-parameter update cannot (Table~\ref{tab:selector}, Figure~\ref{fig:pareto}). MAST removes eight MATH forget examples (a 5.333 point drop) while leaving GSM8K slightly higher than the source and MATH retain within half a point. The same-strength full-parameter update reaches a larger forget drop only by collapsing both retain and GSM8K. The gradient diagnostics match this contrast: MAST has a small delta norm (0.211) and modest forget-loss alignment (0.053), while the full-parameter update has a much larger norm (0.747) and higher alignment (0.076), consistent with a broad, high-collateral move.

Paired bootstrap over the evaluation splits (10{,}000 resamples, resampling questions and comparing the same items before and after unlearning) supports this profile. For MAST the forget drop is significant: the paired-bootstrap 95\% CI $[+2.000, +9.333]$ pp excludes zero, and an exact McNemar test gives $p = 0.0078$ (8 lost, 0 recovered). The collateral drops, by contrast, are statistically indistinguishable from zero (retain 95\% CI $[-4.000, +5.000]$ pp; GSM8K 95\% CI $[-3.125, +1.562]$ pp). For the full-parameter update all three drops exclude zero (forget $[+8.0, +22.7]$, retain $[+3.5, +17.5]$, GSM8K $[+2.3, +10.5]$ pp). A direct paired contrast confirms the tradeoff: the full update's \emph{extra} collateral over MAST is significant on both retain ($+10.0$ pp, 95\% CI $[+4.0, +16.0]$) and GSM8K ($+7.0$ pp, 95\% CI $[+2.7, +11.3]$). The intervention removes target reasoning at near-zero collateral, whereas the full-parameter update reaches more forgetting only by sacrificing over ten points of MATH retain and six of GSM8K. Wherever full-parameter GA forgets meaningfully, its collateral exceeds MAST's at matched forgetting, on both models and both collateral axes (Figure~\ref{fig:pareto}); and the frontier extends to a stronger operating point that removes 45 to 28 MATH forget (17 examples) at only $+2.3$ pp GSM8K and $+3.0$ pp retain. Beyond the forget/retain splits, a zero-shot Massive Multitask Language Understanding (MMLU) \cite{hendrycks2021measuring} check over 1000 examples shows broad capability is preserved: RLVR source 38.1\%, MAST (GA) 37.7\%, and full GA 37.6\%; MAST stays within 0.4 pp of the source.

\begin{table}[t]
\centering
\scriptsize
\begin{tabular*}{\textwidth}{@{\extracolsep{\fill}}lrrrrrr@{}}
\toprule
Method & GSM8K & Forget & Retain & F drop & R drop & G drop \\
\midrule
RLVR source & 188/256 & 45/150 & 77/200 & --- & --- & --- \\
MAST (GA) & 190/256 & 37/150 & 76/200 & 5.333 & 0.500 & -0.781 \\
Full-parameter GA & 172/256 & 22/150 & 56/200 & 15.333 & 10.500 & 6.250 \\
\bottomrule
\end{tabular*}
\caption{Selected MAST (GA) versus the full-parameter boundary on the RLVR checkpoint (50 steps, $\eta=5{\times}10^{-6}$). F/R/G drops are forget, retain, and GSM8K drops in percentage points relative to the RLVR source; negative drops denote accuracy improvements after unlearning.}
\label{tab:selector}
\end{table}

\subsection{Ranking Specificity}
\label{sec:ranking}

\begin{figure}[!htbp]
\centering
\includegraphics[width=\linewidth]{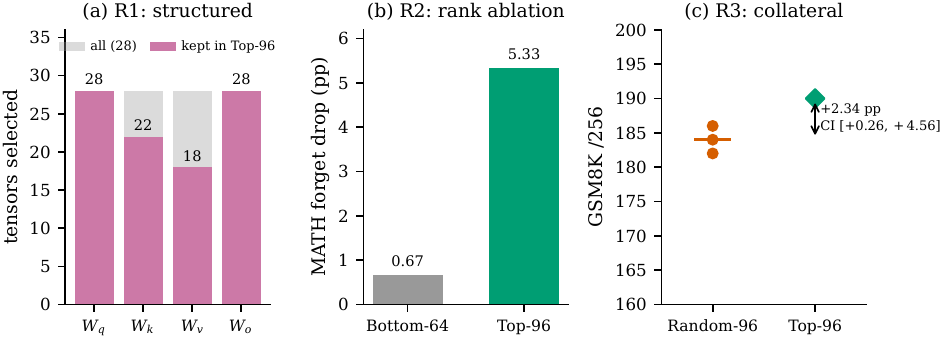}
\caption{Ranking specificity. (a) R1: the top-96 selection is structurally coherent by construction, keeping all query and output projections and dropping only key/value tensors. (b) R2: the bottom-ranked 64 tensors produce negligible forgetting at the selected strength, while the ranked top-96 forgets. (c) R3: at matched forgetting, the ranked selector retains significantly more GSM8K than a same-size random subset (paired bootstrap $+2.34$ pp, 95\% CI excludes zero).}
\label{fig:ranking}
\end{figure}

MAST's advantage over an undifferentiated parameter reduction comes from \emph{which} tensors it updates. Three properties establish that the mechanism-based ranking, not merely the smaller parameter count, drives the result (Figure~\ref{fig:ranking}); together they place MAST in the same specificity paradigm as localization-based editing methods \cite{rome,memit,localization_editing}.

\paragraph{(R1) The selected subset is structurally coherent.} The top-96 operating point keeps all 28 query-projection tensors, all 28 output-projection tensors, 22 key-projection tensors, and 18 value-projection tensors; the 16 excluded tensors are 10 value and 6 key tensors, while all query and output projections are retained across all 28 layers. The ranking selects a structured subset that differs from random selection by construction, not by chance. This composition has a plausible mechanistic reading: query and output projections govern attention routing and the mixing of attended values---the components most directly reshaped when RLVR redistributes probability over reasoning trajectories rather than boosting token likelihood uniformly---so the score of Eq.~\ref{eq:score} ranks them highest.

\paragraph{(R2) The ranking localizes the forgetting.} Updating the bottom-ranked 64 tensors at the same strength yields no meaningful forgetting (MATH forget $45\to44$/150), while the top-ranked subset forgets; because the bottom-64 and top-96 subsets share only 48 mid-rank tensors, the rank ordinal---not the parameter count---carries the forget signal.

\paragraph{(R3) The ranking lowers collateral at matched forgetting.} At a matched forget level, the ranked top-96 subset (GSM8K 190/256) preserves GSM8K significantly better than a same-size random-96 subset (GSM8K 182--186/256): paired bootstrap $+2.34$ pp, 95\% CI $[+0.26, +4.56]$, excluding zero.

\FloatBarrier

\subsection{Robustness Across Strengths, Seeds, and Objectives}

The selected point sits on a tunable strength frontier. Lowering the learning rate to $\eta=3{\times}10^{-6}$ weakens forgetting (MATH forget 41/150, a 2.667 point drop), while raising it to $\eta=1{\times}10^{-5}$ trades collateral for stronger forgetting (MATH forget 28/150, GSM8K 182/256, retain 71/200). The selected operating point reproduces across training seeds: over three seeds the MATH forget count is $\{37, 37, 41\}/150$ (mean 38.3, range $[37, 41]$), and the non-GA NPO variant is similarly stable at $\{41, 39, 39\}/150$ (mean 39.7), so the result is not a single-seed artifact. Forgetting is tunable by learning rate, and a width band reaches the operating point with the selected width cleanest: at the same strength, top-32 gives 44/150, top-88 gives 39/150, top-96 gives 37/150, and top-104 gives 39/150. Adding the next eight lower-ranked tensors (ranks 97--104) reduces forgetting ($37\to39$) at higher collateral (GSM8K $190\to188$), so which tensors are updated---not how many---governs the tradeoff.

The intervention is not specific to gradient ascent. We test three SFT-era unlearning objectives at matched strength (50 steps, $\eta=5{\times}10^{-6}$): gradient ascent (GA), negative preference optimization (NPO, $\beta=0.5$), and SimNPO ($\beta=2.5$, reference-free and length-normalized). In every case the full-parameter update lands on the collateral boundary while MAST reaches meaningful forgetting at much lower collateral (Table~\ref{tab:baselines}). Across three NPO training seeds MAST has mean forget drop 3.556 points, mean retain drop -0.667 points, and mean GSM8K drop 0.000 points, so the restricted-versus-full contrast holds under all three objectives.

\begin{table}[t]
\centering
\scriptsize
\begin{tabular*}{\textwidth}{@{\extracolsep{\fill}}lrrrrrr@{}}
\toprule
Method & GSM8K & Forget & Retain & F drop & R drop & G drop \\
\midrule
RLVR source & 188/256 & 45/150 & 77/200 & --- & --- & --- \\
MAST (GA) & 190/256 & 37/150 & 76/200 & 5.333 & 0.500 & -0.781 \\
Full-parameter GA & 172/256 & 22/150 & 56/200 & 15.333 & 10.500 & 6.250 \\
MAST (NPO) & 188/256 & 39/150 & 80/200 & 4.000 & -1.500 & 0.000 \\
Full-parameter NPO & 168/256 & 23/150 & 54/200 & 14.667 & 11.500 & 7.813 \\
MAST (SimNPO) & 182/256 & 38/150 & 75/200 & 4.667 & 1.000 & 2.344 \\
Full-parameter SimNPO & 172/256 & 20/150 & 54/200 & 16.667 & 11.500 & 6.250 \\
\bottomrule
\end{tabular*}
\caption{Three SFT-era objectives on the RLVR checkpoint (50 steps, $\eta=5{\times}10^{-6}$). For all of GA, NPO, and SimNPO the full-parameter update lands on the collateral boundary while MAST forgets at much lower collateral. F/R/G drops are forget, retain, and GSM8K drops in percentage points; negative drops denote accuracy improvements after unlearning.}
\label{tab:baselines}
\end{table}

\section{Evaluation: Forget Accuracy Alone Is Misleading}
\label{sec:eval}

\begin{figure}[!htbp]
\centering
\includegraphics[width=0.75\linewidth]{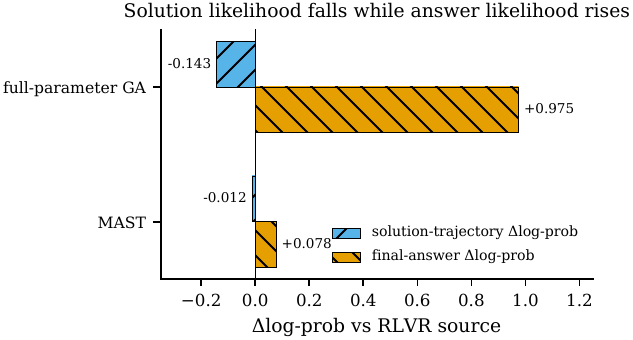}
\caption{Why forget-accuracy alone is misleading. Both the restricted and the full update lower the solution-trajectory log-probability while \emph{raising} the short final-answer log-probability, relative to the RLVR source. Generated-correctness and answer-string-likelihood views therefore disagree.}
\label{fig:eval}
\end{figure}

Our matched setup also exposes a general evaluation pitfall for reasoning unlearning. Mechanism diagnostics on the unlearned models show why forget accuracy alone is insufficient. For each forget example we compute, under the unlearned model relative to the source, two token-averaged log-probability changes: the \emph{solution-trajectory} delta-log-probability (averaged over the full chain-of-thought solution tokens) and the \emph{final-answer} delta-log-probability (averaged over the short boxed-answer tokens). Both the restricted and the full updates lower the solution-trajectory log-probability while raising the final-answer log-probability. The selected restricted GA point has solution delta-log-probability -0.0117 and final-answer delta-log-probability +0.078; the full-parameter GA boundary has solution delta-log-probability -0.143 and final-answer delta-log-probability +0.975. In a per-example case study of the selected point, all eight source-correct examples that flip to wrong have negative solution delta-log-probability and non-negative final-answer delta-log-probability. Generated correctness drops while the answer-string likelihood does not (Figure~\ref{fig:eval}), so answer-likelihood-only and forget-accuracy-only views can disagree; both are incomplete unlearning evidence.

\FloatBarrier

\section{Cross-Model Validation (Qwen3-1.7B-Base)}
\label{sec:crossmodel}

To test whether the findings are specific to one model family, we rerun the entire pipeline---matched checkpoints, mechanism diagnostics, and unlearning---on Qwen3-1.7B-Base, a non-gated, cross-generation base model whose 28-layer / 112-attention-tensor structure is isomorphic to the primary model, so the selector is structurally applicable without redefinition. The cross-model run \emph{corroborates} the three core findings at a per-model calibrated strength.

\subsection{Matched Controls and Mechanism}

The matched checkpoints are within the 3 percentage point threshold, and the mechanism separation reproduces in the same direction, weaker in magnitude (Table~\ref{tab:qwen3-matched}). The SFT--RLVR GSM8K gap is 2.34 percentage points. Mechanism: SFT mean delta-log-probability $+0.257$ / positive-delta fraction 0.571; RLVR increment $+0.001$ / 0.406---the same SFT-positive-boost versus RLVR-balanced structure as the primary model.

\begin{table}[t]
\centering
\small
\begin{tabular*}{\textwidth}{@{\extracolsep{\fill}}lrrr@{}}
\toprule
Qwen3 checkpoint & GSM8K & MATH forget & MATH retain \\
\midrule
Base & 161/256 & 31/150 & 65/200 \\
SFT & 177/256 & 42/150 & 76/200 \\
RLVR & 171/256 & 41/150 & 74/200 \\
\bottomrule
\end{tabular*}
\caption{Matched checkpoint control table (Qwen3-1.7B-Base). SFT and RLVR differ by 2.34 percentage points on GSM8K, below the 3 percentage point gate.}
\label{tab:qwen3-matched}
\end{table}
\FloatBarrier

\subsection{Full-Parameter Boundary and Restricted-versus-Full Collateral Advantage}

The full-parameter collateral boundary reproduces, and the headline cross-model result is that target restriction protects GSM8K accuracy and substantially reduces retain collateral where the full update destroys general performance (Table~\ref{tab:qwen3-unlearn}). Two points are statistically supported by paired bootstrap (B=10{,}000). First, the full-parameter update inflicts significant collateral (GSM8K $+10.9$ pp, 95\% CI $[+4.3, +17.6]$; retain $+14.0$ pp, $[+7.5, +20.5]$). Second, the restricted-versus-full GSM8K collateral advantage is significant: $+12.9$ pp, 95\% CI $[+6.6, +19.1]$, excluding zero---MAST preserves GSM8K accuracy ($+2.0$ pp above source) while the full update sheds nearly eleven points (Figure~\ref{fig:pareto}, right column). This mirrors the primary model's significant collateral advantage with an independently significant cross-model effect.

\begin{table}[!htbp]
\centering
\small
\begin{tabular*}{\textwidth}{@{\extracolsep{\fill}}lrrr@{}}
\toprule
Qwen3 method & GSM8K & MATH forget & MATH retain \\
\midrule
RLVR source & 171/256 & 41/150 & 74/200 \\
Full-parameter GA ($\eta=5{\times}10^{-6}$) & 143/256 ($-10.9$) & 33/150 & 46/200 ($-14.0$) \\
Full-parameter GA ($\eta=8{\times}10^{-6}$) & 40/256 ($-51.2$) & 13/150 & 19/200 ($-27.5$) \\
MAST ($\eta=8{\times}10^{-6}$, selected) & 176/256 ($+2.0$) & 32/150 & 57/200 ($-8.5$) \\
\bottomrule
\end{tabular*}
\caption{Unlearning on Qwen3-1.7B-Base. The restricted-versus-full GSM8K collateral advantage is significant ($+12.9$ pp, 95\% CI $[+6.6, +19.1]$). Parenthetical values are signed changes from the RLVR source in percentage points (negative denotes a decrease, i.e.\ collateral damage); bootstrap CIs in the text are reported on the corresponding drop magnitudes (a positive drop = a decrease).}
\label{tab:qwen3-unlearn}
\end{table}
\FloatBarrier

MAST forgets while preserving GSM8K accuracy on Qwen3 and cutting retain collateral relative to full GA: MATH forget $41\to32$, GSM8K $+2.0$ pp above source, and retain drop 8.5 pp versus the full update's 14.0 pp. The contrast is sharpest at \emph{identical} strength: at $\eta=8{\times}10^{-6}$, full-parameter GA collapses (GSM8K 40/256) while MAST remains a targeted operating point (GSM8K 176/256, MATH forget $41\to32$)---the same update strength destroys the full model yet permits clean forgetting under target restriction (Table~\ref{tab:qwen3-unlearn}).

\section{Reproducibility}

All checkpoints are trained and evaluated with the same scripts and data splits. SFT uses standard cross-entropy on the matched GSM8K data; the RLVR recipe uses Group Relative Policy Optimization (GRPO) \cite{deepseekmath} with full-precision (FP32) full-parameter optimization, $\eta=5{\times}10^{-7}$, gradient-norm clipping at 0.5, left-padded tokenization, and sampling temperature 0.7 / top-$p$ 0.95. Unlearning uses 50 optimizer steps at the per-model calibrated learning rate. Evaluation is greedy exact-match with the tolerant regrade protocol described in Section~\ref{sec:forget-retain}; the MMLU check uses zero-shot log-probability scoring over A/B/C/D on a fixed 1000-example subset. Code, configurations, target-tensor manifests, and per-sample evaluation outputs are released with the paper to support reproduction.

\section{Limitations}

This study covers mathematical reasoning on two model families, with the unlearning strength calibrated per model ($\eta=5{\times}10^{-6}$ primary, $\eta=8{\times}10^{-6}$ Qwen3). Extending MAST to further domains and unlearning objectives is natural future work.

\section{Conclusion}

Across two model families, matched SFT/RLVR controls show that the SFT-to-RLVR increment differs from the SFT update in token-level probability structure, and that standard full-parameter unlearning has no clean operating point on the RLVR checkpoint---it forgets only by inflicting large collateral damage. MAST turns this diagnosis into a method: ranking attention-projection tensors by a mechanism score and restricting the update to the top-ranked subset, it forgets meaningfully at near-zero collateral on the primary model (McNemar $p=0.0078$) and yields a statistically significant collateral advantage over full-parameter unlearning on both models. Our diagnostics also expose a general evaluation pitfall: updates lower solution-trajectory likelihood while raising final-answer likelihood, so forget-accuracy-only and answer-likelihood-only views are each incomplete.

\bibliographystyle{splncs04}
\bibliography{references}

\end{document}